\documentclass{article} 
\usepackage{iclr2020_conference,times}

\usepackage{amsmath,amsfonts,bm}

\def\eqref#1{equation~\ref{#1}}

\def\1{\bm{1}}

\DeclareMathAlphabet{\mathsfit}{\encodingdefault}{\sfdefault}{m}{sl}
\SetMathAlphabet{\mathsfit}{bold}{\encodingdefault}{\sfdefault}{bx}{n}

\newcommand{\E}{\mathbb{E}}

\usepackage[colorlinks=true,citecolor=black,linkcolor=magenta]{hyperref}
\usepackage{url}
\usepackage{array,multirow,graphicx}
\usepackage{booktabs}

\usepackage{adjustbox}

\usepackage{enumitem}
\usepackage{wrapfig}
\usepackage{subcaption}

\title{Outlier Exposure with Confidence Control \\ for Out-of-Distribution Detection}

\iclrfinalcopy

\author{Aristotelis-Angelos Papadopoulos \& Mohammad Reza Rajati \& Nazim Shaikh \& Jiamian Wang\\
University of Southern California\\
Los Angeles, CA 90089, USA \\
\texttt{\{aristop,rajati,nshaikh,jiamianw\}@usc.edu} \\
}

\begin{document}

\maketitle

\begin{abstract}
Deep neural networks have achieved great success in classification tasks during the last years. However, one major problem to the path towards artificial intelligence is the inability of neural networks to accurately detect samples from novel class distributions and therefore, most of the existent classification algorithms assume that all classes are known prior to the training stage. In this work, we propose a methodology for training a neural network that allows it to efficiently detect out-of-distribution (OOD) examples without compromising much of its classification accuracy on the test examples from known classes. We propose a novel loss function that gives rise to a novel method, Outlier Exposure with Confidence Control (OECC), which achieves superior results in OOD detection with OE both on image and text classification tasks without requiring access to OOD samples. Additionally, we experimentally show that the combination of OECC with state-of-the-art post-training OOD detection methods, like the Mahalanobis Detector (MD) and the Gramian Matrices (GM) methods, further improves their performance in the OOD detection task, demonstrating the potential of combining training and post-training methods for OOD detection.\footnote{Our code is publicly available at \href{https://github.com/nazim1021/OOD-detection-using-OECC}{https://github.com/nazim1021/OOD-detection-using-OECC}.}
\end{abstract}

\section{Introduction}
Modern neural networks have recently achieved superior results in classification problems \citep{NIPS2012_4824, DBLP:journals/corr/HeZRS15}. However, most classification algorithms proposed so far assume that samples from all class conditional distributions are available during training time i.e., they make the closed-world assumption. In an open world environment \citep{Bendale_2015_CVPR}, where examples from novel class distributions might appear during test time, it is necessary to build classifiers that are able to detect OOD examples while having high classification accuracy on known class distributions.

It is generally known that deep neural networks can make predictions for out-of-distribution (OOD) examples with high confidence \citep{DBLP:conf/cvpr/NguyenYC15}. High confidence predictions are undesirable since they consist a symptom of overfitting \citep{DBLP:journals/corr/SzegedyVISW15}. They also make the calibration of neural networks difficult. \cite{Guo:2017:CMN:3305381.3305518} observed that modern neural networks are miscalibrated since their average confidence is usually much higher than their accuracy. 

A simple yet effective method to address the problem of the inability of neural networks to detect OOD examples is to train them so that they make highly uncertain predictions for examples generated by novel class distributions. In order to achieve that, \cite{2017arXiv171109325L} defined a loss function based on the Kullback-Leibler (KL) divergence to minimize the distance between the output distribution given by softmax and the uniform distribution for samples generated by a GAN \citep{Goodfellow:2014:GAN:2969033.2969125}. Using a similar loss function, \citet{hendrycks2019oe} showed that the technique of Outlier Exposure (OE) that draws anomalies from a real and diverse dataset can outperform the GAN framework for OOD detection. 

In this paper, based on the idea of \citep{hendrycks2019oe}, our main contribution is threefold:
\begin{itemize}
  \item We propose a novel method, Outlier Exposure with Confidence Control (OECC), consisting of two regularization terms. The first regularization term minimizes the total variation distance between the output distribution given by softmax for an auxiliary dataset and the uniform distribution, which constitutes a distance metric between the two distributions \citep{Gibbs02onchoosing, Deza.Deza2009EncyclopediaofDistances}. The second regularization term minimizes the Euclidean distance between the training accuracy of a DNN and the average confidence in its predictions on the training set.
  
  \item We experimentally show that OECC achieves superior results in OOD detection with OE without requiring access to OOD samples. Additionally, similar to \citep{hendrycks2019oe} and in contrast with many other state-of-the-art OOD detection methods such as the Mahalanobis Detector (MD) \citep{Lee:2018:SUF:3327757.3327819}, the Gramian Matrices (GM) method \citep{ch2019detecting}, ODIN \citep{2017arXiv170602690L} and Res-Flow \citep{Zisselman_2020_CVPR}, we show that OECC can be applied to both image and text classification tasks. Furthermore, the experimental results demonstrate that OECC can successfully detect OOD samples both when in- and out-of-distribution examples are far from each other as well as in the more challenging case when in- and out-of-distribution examples are close.
  
  \item We experimentally show that OECC can be combined with state-of-the-art post-training methods for OOD detection like the Mahalanobis Detector (MD) \citep{Lee:2018:SUF:3327757.3327819} and the Gramian Matrices method (GM) \citep{ch2019detecting}. The experimental results demonstrate that the resulting combination achieves superior results in OOD detection, demonstrating the potential of combining training and post-training
  methods for OOD detection in the future research efforts. 
\end{itemize}

\section{Related Work}
\citet{2017arXiv170508722Y} used the GAN framework \citep{Goodfellow:2014:GAN:2969033.2969125} to generate negative instances of seen classes by finding data points that are close to the training instances but are classified as fake by the discriminator. Then, they used those samples in order to train SVM classifiers to detect examples from unseen classes. Similarly, \citet{2018arXiv180210560K} used a multi-class GAN framework in order to produce a generator that generates a mixture of nominal data and novel data and a discriminator that performs simultaneous classification and novelty detection.

\citet{hendrycks17baseline} proposed a baseline for detecting misclassified and out-of-distibution examples based on their observation that the prediction probability of out-of-distribution examples tends to be lower than the prediction probability for correct examples. \citet{2017arXiv171109325L} generated GAN examples and forced the DNN to have lower confidence in predicting the classes for those examples while in \citet{hendrycks2019oe}, the GAN samples were substituted with a real and diverse dataset using the technique of OE. Similar works \citep{Malinin:2018:PUE:3327757.3327808, 2018arXiv180807703B} also force the model to make uncertain predictions for OOD examples. Using an ensemble of classifiers, \citet{Lakshminarayanan:2017:SSP:3295222.3295387} showed that their method was able to express higher uncertainty in OOD examples. \citet{Hein_2019_CVPR} showed that RELU networks might produce high confidence predictions far away from the training data while in \citet{Meinke2020Towards}, the authors proposed to modify the network architecture by integrating a generative model and they showed that the resulting architecture produces close to uniform predictions far away from the training data. \citet{2018arXiv180800529L} provided theoretical guarantees for detecting OOD examples under the assumption that an upper bound of the fraction of OOD examples is available. Recently, an adversarial training approach that can significantly increase the robustness of OOD detectors was proposed \citep{2020arXiv200615207C}. 

For image data, based on the idea of \cite{hendrycks17baseline}, it was observed that simultaneous use of temperature scaling \citep{Guo:2017:CMN:3305381.3305518} and small perturbations at the input, a method called ODIN \citep{2017arXiv170602690L}, can push the softmax scores of in- and out-of-distribution images further apart from each other, making the OOD images distinguishable. Recently, the ODIN method was generalized to not require access to OOD examples to tune its parameters \citep{generalized_odin}. Under the assumption that the pre-trained features of a softmax neural classifier can be fitted well by a class-conditional Gaussian distribution, one can define a confidence score using the Mahalanobis distance that can efficiently detect abnormal test samples \citep{Lee:2018:SUF:3327757.3327819}. \citet{ch2019detecting} proposed the use of higher order Gram matrices to compute pairwise feature correlations between the channels of each layer of a DNN. The methods proposed by \citet{Lee:2018:SUF:3327757.3327819}, \citet{ch2019detecting} and \citet{2017arXiv170602690L} are post-training methods for OOD detection. 

Recently, there is also a growing interest in applying machine learning in a self-supervised manner for OOD detection. \citet{hendrycks2019using} combined different self-supervised geometric translation prediction tasks in one model, using multiple auxiliary heads. They showed that their method performs well on detecting outliers which are close to the in-distribution data. \citet{mohseni2020} proposed using one auxiliary head in a self-supervised manner to learn generalizable OOD features.

Unsupervised methods have also been studied for OOD detection. \citet{choi2018waic} showed that in high dimensions, likelihood models might assign high likelihoods to OOD inputs. \citet{2018arXiv181009136N} experimentally showed that deep generative models assign higher likelihood to OOD data compared to in-distribution data. To mitigate this problem, \cite{2019arXiv190602845R} proposed a likelihood ratio method that improves the OOD detection capabilities of deep generative models of images. Recently, \cite{morningstar2020density} and \cite{erdil2020unsupervised} proposed unsupervised methods for OOD detection based on kernel density estimation.

\section{Outlier Exposure with Confidence Control (OECC)}\label{teleftaio}
We consider the multi-class classification problem under the open-world assumption \citep{Bendale_2015_CVPR}, where samples from some classes are not available during training. Our task is to design deep neural network classifiers that can achieve high accuracy on examples generated by a learned probability distribution called $D_{in}$ while at the same time, they can effectively detect examples generated by a different probability distribution called $D_{out}$ during the test phase. The examples generated by $D_{in}$ are called in-distribution while the examples generated by $D_{out}$ are called out-of-distribution (OOD). Adopting the idea of Outlier Exposure (OE) \citep{hendrycks2019oe}, we train the neural network using training examples sampled from $D_{in}$ and $D^{OE}_{out}$. During the test phase, we evaluate the OOD detection capability of the neural network using examples sampled from $D^{test}_{out}$, where $D^{OE}_{out}$ and $D^{test}_{out}$ are disjoint.

In previous works \citep{2017arXiv171109325L, hendrycks2019oe}, the KL divergence metric was used in order to minimize the distance between the output distribution produced by softmax for the OOD examples and the uniform distribution. However, it is generally known that KL divergence does not satisfy the symmetry and the triangle inequality properties as required by a distance metric \citep{Gibbs02onchoosing,Deza.Deza2009EncyclopediaofDistances}. 
In our work, we choose to minimize the total variation distance \citep{Gibbs02onchoosing} between the two distributions. There are several reasons for this choice. First, the total variation distance satisfies all the properties required by a distance metric. Second, it is one of the most commonly
used probability metrics since it admits natural interpretations. For instance, in Bayesian statistics, the error in a bounded expected loss function due to the approximation of a probability measure by another is given by the total
variation distance \citep{Gibbs02onchoosing}. Moreover, as we also mention later, the total variation distance has the unique property to uniformly attract all the prediction probabilities produced by softmax for data sampled from $D_{out}^{OE}$ towards the uniform distribution, making the neural network better detect in- and out-of-distribution examples.

Viewing the knowledge of a model as the class conditional distribution it produces over outputs given an input \citep{HinVin15Distilling}, the entropy of this conditional distribution can be used as a regularization method that penalizes confident predictions of a neural network \citep{DBLP:conf/iclr/PereyraTCKH17}. In our approach, instead of penalizing the confident predictions of posterior probabilities yielded by a neural network, we force it to make predictions for examples generated by $D_{in}$ with an average confidence close to its training accuracy. In such a manner, not only do we make the neural network avoid making overconfident predictions, but we also take into consideration its calibration \citep{Guo:2017:CMN:3305381.3305518}.

Let us consider a classification model that can be represented by a parametrized function $f_{\boldsymbol{\theta}}$, where $\boldsymbol{\theta}$ stands for the vector of parameters in $f_{\boldsymbol{\theta}}$. Without loss of generality, assume that the cross-entropy loss function is used during training. We propose the following constrained optimization problem for finding $\boldsymbol{\theta}$:
\begin{equation}
\label{optim1}
\begin{aligned}
& \underset{\boldsymbol{\theta}}{\text{minimize}}
& & \E_{(x,y)\sim D_{in}}[\mathcal{L}_{CE}(f_{\boldsymbol{\theta}}(x),y)] \\
& \text{subject to}
& & \E_{x\sim D_{in}} \Bigg[ \underset{l=1,...,K} \max\Bigg(\frac{e^{z_l}}{\sum_{j=1}^K e^{z_j}}\Bigg) \Bigg] = A_{tr} \\
&&& \underset{l=1,...,K} \max\Bigg(\frac{e^{z_l}}{\sum_{j=1}^K e^{z_j}}\Bigg) = \frac{1}{K}, \; \forall x^{(i)} \sim D^{OE}_{out}
\end{aligned}
\end{equation}

where $\mathcal{L}_{CE}$ is the cross-entropy loss function and $K$ is the number of classes available in $D_{in}$. Even though the constrained optimization problem (\ref{optim1}) can be used for training various classification models, for clarity we limit our discussion to deep neural networks. Let $\bf z$ denote the vector representation of the example $x^{(i)}$ in the feature space produced by the last layer of the deep neural network (DNN) and let $A_{tr}$ be the training accuracy of the DNN. Observe that the optimization problem (\ref{optim1}) minimizes the cross entropy loss function subject to two additional constraints. The first constraint forces the average maximum prediction probabilities calculated by the softmax layer towards the training accuracy of the DNN for examples sampled from $D_{in}$, while the second constraint forces the maximum probability calculated by the softmax layer towards $\frac{1}{K}$ for all examples sampled from the probability distribution $D^{OE}_{out}$. In other words, the first constraint makes the DNN predict examples from known classes with an average confidence close to its training accuracy, while the second constraint forces the DNN to be highly uncertain for examples of classes it has never seen before by producing a uniform distribution at the output for examples sampled from the probability distribution $D^{OE}_{out}$. It is also worth noting that the first constraint of (\ref{optim1}) uses the training accuracy of the neural network $A_{tr}$ which is not available in general. To handle this issue, one can train a neural network by only minimizing the cross-entropy loss function for a few number of epochs in order to estimate $A_{tr}$ and then fine-tune it using (\ref{optim1}). 

\par Because solving the nonconvex constrained optimization problem described by (\ref{optim1}) is extremely difficult, let us introduce Lagrange multipliers \citep{citeulike:163662} and convert (\ref{optim1}) into the following unconstrained optimization problem:
\begin{equation}
\label{optim2}
\begin{aligned}
\underset{\boldsymbol{\theta}}{\text{minimize}}\hspace{5pt} \E_{(x,y)\sim D_{in}}&[\mathcal{L}_{CE}(f_{\boldsymbol{\theta}}(x),y)] \\&+ \lambda_1 \Bigg(A_{tr}-\E_{x\sim D_{in}} \Bigg[ \underset{l=1,...,K} \max\Bigg(\frac{e^{z_l}}{\sum_{j=1}^K e^{z_j}}\Bigg) \Bigg]\Bigg) \\&+  \lambda_2 \sum_{x^{(i)} \sim D^{OE}_{out}} \Bigg(\frac{1}{K}-\underset{l=1,...,K} \max\Bigg(\frac{e^{z_l}}{\sum_{j=1}^K e^{z_j}}\Bigg)\Bigg)
\end{aligned}
\end{equation}
where it is worth mentioning that in (\ref{optim2}), we used only one Lagrange multiplier for the second set of constraints in (\ref{optim1}) instead of using one for each constraint in order to avoid introducing a large number of hyperparameters to our loss function. This modification is a special case where we consider the Lagrange multiplier $\lambda_2$ to be common for each individual constraint involving a different $x^{(i)} \sim D^{OE}_{out}$. Note also that according to the original Lagrangian theory, one should optimize the objective function of (\ref{optim2}) both with respect to $\boldsymbol{\theta}$, $\lambda_1$ and $\lambda_2$ but as it is common in machine learning applications, we approximate the original problem by calculating appropriate values for $\lambda_1$ and $\lambda_2$ through a validation technique \citep{hastie01statisticallearning}.  

After converting the constrained optimization problem (\ref{optim1}) into an unconstrained optimization problem described by (\ref{optim2}), we observed in the simulation experiments that at each training epoch, the maximum prediction probability produced by softmax for each example drawn from $D^{OE}_{out}$ changes, introducing difficulties in making the DNN produce a uniform distribution at the output for those examples. For instance, assume that we have a $K$-class classifier with $K=3$ and at epoch $t_n$, the maximum prediction probability produced by softmax for an example $x^{(i)} \sim D^{OE}_{out}$ corresponds to the second class. Then, the last term of (\ref{optim2}) will push the prediction probability of example $x^{(i)}$ for the second class towards $\frac{1}{3}$ while concurrently increasing the prediction probabilities for either the first class or the third class or both. At the next epoch $t_{n+1}$, it is possible that the prediction probability for either the first class or the third class becomes the maximum among the three and hence, the last term of (\ref{optim2}) will push that one towards $\frac{1}{3}$ by possibly increasing again the prediction probability for the second class. It becomes obvious that this process introduces difficulties in making the DNN produce a uniform distribution at the output for examples sampled from $D^{OE}_{out}$. Fortunately, this issue can be resolved by concurrently pushing all the prediction probabilities produced by the softmax layer for examples drawn from $D^{OE}_{out}$ towards $\frac{1}{K}$.    

Additionally, in order to prevent the second and the third term of (\ref{optim2}) from taking negative values during training, let us convert (\ref{optim2}) into the following:  
\begin{equation}
\label{optim3}
\begin{aligned}
\underset{\boldsymbol{\theta}}{\text{minimize}}\hspace{5pt} \mathcal\E_{(x,y)\sim D_{in}}&[\mathcal{L}_{CE}(f_{\boldsymbol{\theta}}(x),y)] \\& + \lambda_1 \Bigg(A_{tr}-\E_{x\sim D_{in}} \Bigg[ \underset{l=1,\dots,K} \max\Bigg(\frac{e^{z_l}}{\sum_{j=1}^K e^{z_j}}\Bigg) \Bigg]\Bigg)^2 \\&+  \lambda_2 \sum_{x^{(i)} \sim D^{OE}_{out}} \sum_{l=1}^K \Bigg|\frac{1}{K}-\frac{e^{z_l}}{\sum_{j=1}^K e^{z_j}}\Bigg|
\end{aligned}
\end{equation}

The second term of the the loss function described by (\ref{optim3}) minimizes the squared distance between the training accuracy of the DNN and the average confidence in its predictions for examples drawn from $D_{in}$. Additionally, the third term of (\ref{optim3}) minimizes the $l_1$ norm between the uniform distribution and the distribution produced by the softmax layer for the examples drawn from $D^{OE}_{out}$. 

At this point, let $\Omega$ be a countable sample space and let $P, Q$ denote two probability measures on $\Omega$. Then, the total variation distance between $P$ and $Q$ is defined as: 
\begin{equation}
    \label{total_variation}
    d_{TV}(P,Q) = \frac{1}{2} \sum_{x \in \Omega} |P(x)-Q(x)|
\end{equation}
Let $\Omega$ be the space of $D_{out}^{OE}$, $P$ be the discrete uniform distribution and $Q$ be the probability distribution produced by the softmax layer for the examples sampled from $D_{out}^{OE}$. Then, the third term of (\ref{optim3}) is equivalent to the total variation distance described by (\ref{total_variation}). We call the methodology of training a DNN with the loss function described by (\ref{optim3}) Outlier Exposure with Confidence Control (OECC).

While converting the unconstrained optimization problem (\ref{optim2}) into (\ref{optim3}), one could use several combinations of norms for regularization. However, we found that minimizing the squared distance between the training accuracy of the DNN and the average confidence in its predictions for examples drawn from $D_{in}$ and the total variation distance (or equivalently, $l_1$ norm) between the uniform distribution and the distribution produced by the softmax layer for the examples drawn from $D^{OE}_{out}$ works best. This is because $l_1$ norm uniformly attracts all the prediction probabilities produced by softmax to the desired value $\frac{1}{K}$, better contributing to producing a uniform distribution at the output of the DNN for the examples drawn from $D^{OE}_{out}$. On the other hand, minimizing the squared distance between the training accuracy of the DNN and the average confidence in its predictions for examples drawn from $D_{in}$ emphasizes more on attracting the maximum softmax probabilities that are further away from the training accuracy of the DNN, making the neural network better detect in- and out-of-distribution examples at the low softmax probability levels.

\section{Experiments}
During the experiments, we observed that if we start training the DNN with a relatively high value of $\lambda_1$, the learning process might slow down since we constantly force the neural network to make predictions with an average confidence close to its training accuracy, which is initially low mainly due to the random weight initialization. Therefore, it is recommended to split the training of the algorithm into two stages where in the first stage, we train the DNN using only the cross-entropy loss function until it reaches the desired level of accuracy $A_{tr}$ and then using a fixed $A_{tr}$, we fine-tune it using the OECC method.\footnote{Our code is publicly available at \href{https://github.com/nazim1021/OOD-detection-using-OECC}{https://github.com/nazim1021/OOD-detection-using-OECC}.}

Many of the proposed methods for OOD detection can be classified into two categories. In the first category of methods, the DNN is trained in a manner that improves its OOD detection capability. This category of methods includes both supervised \citep{2017arXiv171109325L,hendrycks2019oe} and self-supervised approaches \citep{hendrycks2019using,mohseni2020}. These methods can be called training methods for OOD detection. OECC belongs to the category of training methods. The second category are post-training methods for OOD detection \citep{Lee:2018:SUF:3327757.3327819,ch2019detecting,2017arXiv170602690L, Zisselman_2020_CVPR} where the OOD detection method is applied after training a neural network.

We conducted two types of experiments that are presented in Sections 4.1 and 4.2. In Section 4.1, we compare our method with the Outlier Exposure (OE) method \citep{hendrycks2019oe}, which is a state-of-the-art method for OOD detection. Since OE is a training method for OOD detection, we adopt the experimental setup that is commonly used in the training methods for OOD detection. On the other hand, in Section 4.2, we show how OECC can be effectively combined with post-training methods for OOD detection such as the Mahalanobis Detector (MD) \citep{Lee:2018:SUF:3327757.3327819} and the Gramian Matrices (GM) method \citep{ch2019detecting}. Consequently in Section 4.2, we adopt the experimental setup that is commonly used by the post-training methods for OOD detection. 

\subsection{Comparison with State-of-the-Art in OE}
The experimental setting at this section is as follows. We draw samples from $D_{in}$ and we train the DNN using only the cross-entropy loss function until it reaches the desired level of accuracy $A_{tr}$. Then, drawing samples from $D^{OE}_{out}$, we fine-tune it using the OECC method given by (\ref{optim3}). During the test phase, we evaluate the OOD detection capability of the DNN using examples from $D^{test}_{out}$ which is disjoint from $D^{OE}_{out}$. We demonstrate the effectiveness of our method in both image and text classification tasks by comparing it with the previous OOD detection with OE \citep{hendrycks2019oe}, which is a state-of-the-art method in OE. 

\subsubsection{Evaluation Metrics}\label{similar_metrics}
Our method belongs to the class of training methods for OOD detection. Therefore, for the comparison with the OE method, we adopt the evaluation metrics that were also used in other training methods for OOD detection \citep{hendrycks2019oe,hendrycks2019using,mohseni2020}. Defining the OOD examples as the positive class and the in-distribution examples as the negative class, the performance metrics associated with OOD detection are the following:
\addtolength{\leftskip}{0mm}
  \begin{itemize}[leftmargin=\dimexpr\parindent+0mm+0.5\labelwidth\relax]
  \vspace{-5pt}
  \item False Positive Rate at $N\%$ True Positive Rate ({\it FPRN}):
  This performance metric \citep{BMVC2016_119,2015arXiv151209272G} measures the capability of an OOD detector when the maximum softmax probability threshold is set to a predefined value. More specifically, assuming $N\%$ of OOD examples need to be detected during the test phase, we calculate a threshold in the softmax probability space and given that threshold, we measure the false positive rate, i.e. the ratio of in-distribution examples that are incorrectly classified as OOD.
  \item Area Under the Receiver Operating Characteristic curve (AUROC): In the out-of-distribution detection task, the ROC curve \citep{Davis:2006:RPR:1143844.1143874} summarizes the performance of an OOD detection method for varying threshold values.
  \item Area Under the Precision-Recall curve (AUPR): The AUPR \citep{Manning:1999:FSN:311445} is an important measure when there exists a class-imbalance between OOD and in-distribution examples in a dataset. Similar to \cite{hendrycks2019oe}, in our experiments, the ratio of OOD and in-distribution test examples is 1:5.  
\end{itemize}

\subsubsection{Image Classification Experiments}\label{refer_cal_1}
\paragraph{Results.} The results of the image classification experiments are shown in Table~\ref{Image_OE_zipped_exp}. In Figure~\ref{fig:wrapfig1}, as an example, we plot the histogram of softmax probabilities using CIFAR-10 as $D_{in}$ and Places365 as $D^{test}_{out}$. It can be easily observed that a Maximum Softmax Probability (MSP) detector makes predictions for Places365 samples with very high probability, making it impossible for a threshold-based detector to separate in- and out-of-distribution data. On the other hand, after fine-tuning with the OECC method given by (\ref{optim3}), it can be observed that the DNN is making predictions for Places365 samples with very low confidence, forming a uniform distribution at the output for these samples. Therefore, in this case, one can easily design a threshold-based detector that can separate in- and out-of-distribution data.  

The detailed description of the image datasets used as $D_{in}$, $D_{out}^{OE}$ and $D_{out}^{test}$ in the image OOD detection experiments is presented in \ref{imagedata}. The validation data $D_{out}^{val}$ used to tune the hyperparameters $\lambda_1$ and $\lambda_2$ for image classification experiments are presented in \ref{val_image}. Note that $D_{out}^{val}$ and $D_{out}^{test}$ are disjoint.

\begin{wrapfigure}{r}{0.31\textwidth}
\includegraphics[width=1.0\linewidth]{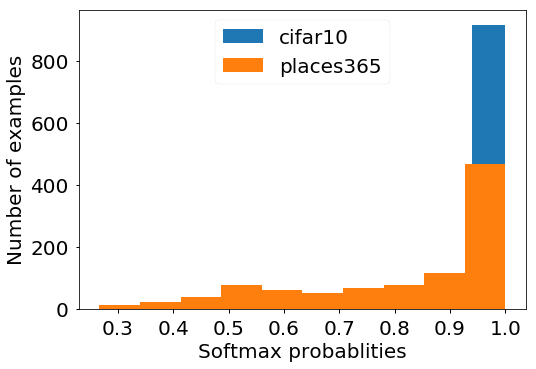} 
\label{fig:wrapfig1}
\includegraphics[width=1.0\linewidth]{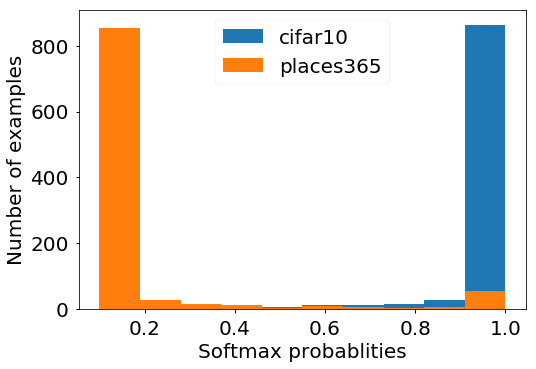} 
\caption{Histogram of softmax probabilities with CIFAR-10 as $D_{in}$ and Places365 as $D^{test}_{out}$ (1,000 samples from each dataset). {\it Top}: MSP baseline detector. {\it Bottom}: MSP detector fine-tuned with
(\ref{optim3}).} 
\label{fig:wrapfig1}
\end{wrapfigure}

\paragraph{Network Architecture and Training Details.}  Similar to \cite{hendrycks2019oe}, CIFAR-10, CIFAR-100 and SVHN datasets were used as $D_{in}$. For CIFAR-10 and CIFAR-100 experiments, we used 40-2 Wide Residual Networks (WRNs) \citep{BMVC2016_87}. We initially trained the WRN for 100 epochs using a cosine learning rate \citep{loshchilov-ICLR17SGDR} with an initial value 0.1, a dropout rate of 0.3 and a batch size of 128. As in \cite{hendrycks2019oe}, we also used Nesterov momentum and $l_2$ weight regularization with a decay factor of 0.0005. For CIFAR-10, we fine-tuned the network for 15 epochs with the OECC method described by (3) using a learning rate of 0.001, while for the CIFAR-100 the corresponding number of epochs was 30. For the SVHN experiments, we trained 16-4 WRNs using a cosine learning rate with an initial value 0.01, a dropout rate of 0.4 and a batch size of 128. We then fine-tuned the network for 5 epochs using a learning rate of 0.001. During fine-tuning, the 80 Million Tiny Images dataset was used as $D^{OE}_{out}$. The values of the hyperparameters $\lambda_1$ and $\lambda_2$ were chosen in the range $[0.03,0.09]$ using a separate validation dataset $D_{out}^{val}$ similar to \cite{hendrycks2019oe}. Note that $D_{out}^{val}$ and $D_{out}^{test}$ are disjoint. The data used for validation are presented in \ref{val_image}.

\begin{table}[t]
\begin{center}
\begin{tabular}{c|cc|cc|cc}
\multicolumn{1}{c}{}&\multicolumn{2}{c}{FPR95$\downarrow$}&\multicolumn{2}{c}{AUROC$\uparrow$}&\multicolumn{2}{c}{AUPR$\uparrow$}\\
\cline{2-7} 
${D}_{in}$&+OE&OECC&+OE&OECC&+OE&OECC\\
\hline
\multirow{1}{*}{{{SVHN}}}&0.10&\textbf{0.03}&99.98&\textbf{99.99}&\textbf{99.83}&99.55\\
\multirow{1}{*}{{{CIFAR-10}}}&9.50&\textbf{6.56}&97.81&\textbf{98.40}&90.48&\textbf{93.08}\\
\multirow{1}{*}{{{CIFAR-100}}}&38.50&\textbf{28.89}&87.89&\textbf{91.80}&58.15&\textbf{71.50}\\
\hline
\end{tabular}
\end{center}
\caption{\label{Image_OE_zipped_exp}Image OOD example detection for the maximum softmax probability (MSP) baseline detector after fine-tuning with OE \citep{hendrycks2019oe} versus fine-tuning with OECC given by (\ref{optim3}). All results are percentages and averaged over 10 runs and over 8 OOD datasets. Detailed experimental results are shown in Appendix~\ref{image_expanded}.}
\end{table}

\paragraph{Contribution of each regularization term.}To demonstrate the effect of each regularization term of the OECC method described by (\ref{optim3}) in the OOD detection task, we ran some additional image classification experiments that are presented in Table~\ref{Ablation_Study}. For these experiments, we incrementally added each regularization term to the loss function described by (\ref{optim3}) and we measured its effect both in the OOD detection evaluation metrics as well as in the accuracy of the DNN on the test images of $D_{in}$. The values of the hyperparameters $\lambda_1$ and $\lambda_2$ were chosen in the range $[0.03,0.09]$ using a separate validation dataset $D_{out}^{val}$ presented in \ref{val_image}. The results of these experiments validate that the combination of the two regularization terms of (\ref{optim3}) not only improves the OOD detection performance of the DNN but also improves its accuracy on the test examples of $D_{in}$ compared to the case where $\lambda_1 = 0$. Table~\ref{Ablation_Study} also demonstrates that our method can significantly improve the OOD detection performance of the DNN compared to the case where only the cross-entropy loss is minimized at the expense of only an insignificant degradation in the test accuracy of the DNN on examples generated by $D_{in}$.    
\begin{table}[h]
\begin{center}
\begin{tabular}{ccccccc}
\cline{1-7} 
$D_{in}$&$\lambda_1$&$\lambda_2$&FPR95$\downarrow$&AUROC$\uparrow$&AUPR$\uparrow$&Test Accuracy($D_{in}$)\\
\hline
\multirow{3}{*}{{{CIFAR-10}}}&-&-&34.94&89.27&59.16&94.65\\
&-&\checkmark&8.87&96.72&77.65&92.72\\
&\checkmark&\checkmark&6.56&98.40&93.08&93.86\\
\cline{1-7}
\multirow{3}{*}{{{CIFAR-100}}}&-&-&62.66&73.11&30.05&75.73\\
&-&\checkmark&26.75&91.59&68.27&71.29\\
&\checkmark&\checkmark&28.89&91.80&71.50&73.14\\
\hline
\end{tabular}
\end{center}
\caption{\label{Ablation_Study}Contribution of each regularization term of (\ref{optim3}) on the OOD detection performance and the test accuracy of the DNN. Results are averaged over 10 runs and over 8 OOD datasets.}
\end{table}

\subsubsection{Text Classification Experiments}\label{refer_cal_2}

\paragraph{Results.} The results of the text classification experiments are shown in Table~\ref{NLP_OE_zipped_exp}. The detailed description of the text datasets used as $D_{in}$, $D_{out}^{OE}$ and $D_{out}^{test}$ in the NLP OOD detection experiments is presented in \ref{nlpdata}.
The validation data $D_{out}^{val}$ used to tune the hyperparameters $\lambda_1$ and $\lambda_2$ for image classification experiments are presented in \ref{val_text}. Note that $D_{out}^{val}$ and $D_{out}^{test}$ are disjoint.
\begin{table}[h]
\begin{center}
\begin{tabular}{c|cc|cc|cc}
\multicolumn{1}{c}{}&\multicolumn{2}{c}{FPR90$\downarrow$}&\multicolumn{2}{c}{AUROC$\uparrow$}&\multicolumn{2}{c}{AUPR$\uparrow$}\\
\cline{2-7} 
${D}_{in}$&+OE&OECC&+OE&OECC&+OE&OECC\\
\hline
\multirow{1}{*}{{{20 Newsgroups}}}&4.86&\textbf{0.63}&97.71&\textbf{99.18}&91.91&\textbf{97.02}\\
\multirow{1}{*}{{{TREC}}}&0.78&\textbf{0.75}&99.28&\textbf{99.32}&\textbf{97.64}&97.52\\
\multirow{1}{*}{{{SST}}}&27.33&\textbf{17.91}&89.27&\textbf{93.79}&59.23&\textbf{74.10}\\
\hline
\end{tabular}
\end{center}
\caption{\label{NLP_OE_zipped_exp}NLP OOD example detection for the maximum softmax probability (MSP) baseline detector after fine-tuning with OE \citep{hendrycks2019oe} versus fine-tuning with OECC given by (\ref{optim3}). All results are percentages and averaged over 10 runs and over 10 OOD datasets. Detailed experimental results are shown in Appendix~\ref{text_expanded}.}
\end{table}
\vspace{-10pt}
\paragraph{Network Architecture and Training Details.} For all text classification experiments, similar to \cite{hendrycks2019oe}, we train 2-layer GRUs \citep{conf/emnlp/ChoMGBBSB14} for 5 epochs with learning rate 0.01 and a batch size of 64 and then we fine-tune them for 2 epochs using the OECC method described by (\ref{optim3}) with a batch size 64 and learning rate 0.01. During fine-tuning, the WikiText-2 dataset was used as $D^{OE}_{out}$. The values of the hyperparameters $\lambda_1$ and $\lambda_2$ were chosen in the range $[0.04,0.1]$ using a separate validation dataset as described in \ref{val_text}.

\paragraph{Discussion.} As can be observed from the results of Table~\ref{Image_OE_zipped_exp} and Table~\ref{NLP_OE_zipped_exp}, OECC achieves superior results in OOD detection compared to the previously proposed OE method \citep{hendrycks2019oe} for both image and text classification tasks. Additionally, the experimental results of Table~\ref{Ablation_Study} suggest that the combination of the two regularization terms of the OECC method is needed in order to maximize the OOD detection capability of the DNN. 
At this point, it is worth mentioning that the experimental results of Table~\ref{Image_OE_zipped_exp} and Table~\ref{NLP_OE_zipped_exp} show that OECC can successfully detect OOD samples in case where in- and out-of-distribution examples are far from each other as well as in the more challenging case where in- and out-of-distribution examples are close. More specifically, \citet{winkens2020contrastive} proposed the confusion log probability (CLP) metric to measure how close (or far) in-distribution and out-of-distribution examples are. In the case where CIFAR-10 is considered as $D_{in}$ and SVHN as $D^{test}_{out}$, $\text{CLP}=[-12.1,-7.6]$ denoting that in- and out-of-distribution data are far from each other. In this case, OECC achieves AUROC 99.6 (see \ref{image_expanded}) which is higher than the one achieved by the joint contrastive \citep{simCLRv1} and supervised learning scheme proposed in \citet{winkens2020contrastive}. On the other hand, in the case where CIFAR-100 is considered as $D_{in}$ and CIFAR-10 as $D^{test}_{out}$, $\text{CLP}=[-4.5,-2.6]$ denoting that in- and out-of-distribution examples are close. In this more challenging case, OECC achieves AUROC 78.7 (see \ref{image_expanded}) which is higher than the one achieved by the joint contrastive and supervised learning, which has been specifically designed for this setting. Note that \citet{winkens2020contrastive} only use the AUROC metric in their results.


\subsection{Combination of OECC and Post-Training methods for OOD Detection}
The experimental setting at this section is as follows. We draw samples from $D_{in}$ and train the DNN until it reaches the desired level of accuracy $A_{tr}$. Then, drawing samples from $D^{OE}_{out}$, we fine-tune it using the OECC method given by (\ref{optim3}). Finally, we apply the post-training method to the fine-tuned model. During the test phase, we evaluate the OOD detection capability of the post-processed model using examples from $D^{test}_{out}$, which is disjoint from $D^{OE}_{out}$.

The detailed description of the image datasets used as $D_{in}$, $D_{out}^{OE}$ and $D_{out}^{test}$ in the image OOD detection experiments is presented in \ref{post_image_data}. The validation data $D_{out}^{val}$ used to tune the hyperparameters $\lambda_1$ and $\lambda_2$ for image classification experiments are presented in \ref{post_val_image}.

\subsubsection{Evaluation Metrics}
To follow the convention in the literature of post-training methods for OOD detection \citep{Lee:2018:SUF:3327757.3327819,ch2019detecting,2017arXiv170602690L, Zisselman_2020_CVPR} and to further demonstrate the adaptability of our method, in the experiments where we combine OECC with post-training methods, we adopt the following OOD detection evaluation metrics. In order to calculate these metrics, we consider in-distribution as positive examples and OOD as negative examples.

\addtolength{\leftskip}{0mm}
  \begin{itemize}[leftmargin=\dimexpr\parindent+0mm+0.5\labelwidth\relax]
  \item True Negative Rate at $N\%$ True Positive Rate ({\it TNRN}):
  This performance metric measures the capability of an OOD detector to detect true negative examples when the true positive rate is set to $95\%$.
  \item Area Under the Receiver Operating Characteristic curve (AUROC): Similar to Section~\ref{similar_metrics}. 
   \item Detection Accuracy (DAcc): This evaluation metric corresponds to the maximum classification accuracy that we can achieve between in- and out-of-distribution examples over all possible thresholds $\epsilon$:
  \begin{equation*}
  \begin{aligned}
  1 - \underset{\epsilon}{\text{min}}\{P_{in}&(q(\boldsymbol{x})\leq \epsilon)P(\boldsymbol{x}\text{ is from } D_{in}) \\&+ P_{out}(q(\boldsymbol{x}) > \epsilon)P(\boldsymbol{x}\text{ is from } D_{out})\},
  \end{aligned}
  \end{equation*}
  where $q(\boldsymbol{x})$ is a confidence score. Similar to \cite{Lee:2018:SUF:3327757.3327819}, we assume that:
  \begin{equation*}
     P(\boldsymbol{x}\text{ is from } D_{in}) = P(\boldsymbol{x}\text{ is from } D_{out}) 
  \end{equation*}
  \item Area Under the Precision Recall curve (AUPR): The PR curve plots the precision against the recall for a varying threshold. In our experiments, we denote by AUPRin (or AUPRout) the area under the PR curve when in- (or out-of-) distribution examples are considered as the positive class.
\end{itemize}

\subsubsection{A Combination of OECC and Mahalanobis Detection Method for OOD Detection}
\citet{Lee:2018:SUF:3327757.3327819} proposed a post-training method for OOD detection that can be applied to any pre-trained softmax neural classifier. Under the assumption that the pre-trained features of a DNN can be fitted well by a class-conditional Gaussian distribution, they defined a confidence score using the Mahalanobis distance with respect to the closest class-conditional probability distribution, where its parameters are chosen as empirical class means and tied empirical covariance of training samples \citep{Lee:2018:SUF:3327757.3327819}. This confidence score can be used as a threshold to determine whether an example is in- or out-of-distribution. To further distinguish in- and out-of-distribution examples, they proposed two additional techniques. In the first technique, they added a small perturbation before processing each input example to increase the confidence score of their method. In the second technique, they proposed a feature ensemble method in order to obtain a better calibrated score. The feature ensemble method extracts all the hidden features of the DNN and computes their empirical class mean and tied covariances. Subsequently, it calculates the Mahalanobis distance-based confidence score for each layer and finally calculates the weighted average of these scores by training a logistic regression detector using validation samples in order to calculate the weight of each layer at the final confidence score.   

Since the Mahalanobis distance-based detector proposed by \citet{Lee:2018:SUF:3327757.3327819} is a post-training method, it can be combined with the proposed OECC method described by (\ref{optim3}). More specifically, in our experiments, we initially trained a DNN using the standard cross-entropy loss function and then we fine-tuned it with the OECC method given by (\ref{optim3}). After fine-tuning, we applied the MD method and we compared the results obtained with the results presented in \citet{Lee:2018:SUF:3327757.3327819}. The simulation experiments on image classification tasks show that the combination of our method which is a training method, with the MD method which is a post-training method achieves state-of-the-art results in the OOD detection task. A part of our experiments was based on the publicly available code of \citet{Lee:2018:SUF:3327757.3327819}.

\paragraph{Experimental Setup.}\label{section_4_2_2}
To demonstrate the adaptability and the effectiveness of our method, we adopt the experimental setup of \citet{Lee:2018:SUF:3327757.3327819}. We train ResNet \citep{DBLP:journals/corr/HeZRS15} with 34 layers using CIFAR-10, CIFAR-100, and SVHN datasets as $D_{in}$. For the CIFAR experiments, SVHN, TinyImageNet (a sample of 10,000 images drawn from the ImageNet dataset) and LSUN are used as $D_{out}^{test}$. For the SVHN experiments, CIFAR-10, TinyImageNet, and LSUN are used as $D_{out}^{test}$. Both TinyImageNet and LSUN images are downsampled to $32\times32$. 

Similar to \citet{Lee:2018:SUF:3327757.3327819}, for the MD method we train the ResNet model for 200 epochs with batch size 128 by minimizing the cross-entropy loss using the SGD algorithm with momentum 0.9. The learning rate starts at 0.1 and is dropped by a factor of 10 at 50\% and 75\% of the training progress, respectively. Subsequently, we compute the Mahalanobis distance-based confidence score using both the input pre-processing and the feature ensemble techniques. The hyper-parameters that need to be tuned are the magnitude of the noise added at each test input example as well as the layer indexes for feature ensemble. Similar to \citet{Lee:2018:SUF:3327757.3327819}, both of them are tuned using a separate validation dataset consisting of both in- and out-of-distribution data since the MD method originally requires access to OOD samples. 

As mentioned earlier, since the Mahalanobis Detector (MD) is a post-training method for OOD detection, it can be combined with our proposed method. More specifically, we initially train the ResNet model with 34 layers for 200 epochs using exactly the same training setup as mentioned above. Subsequently, we fine-tune the network with the OECC method described by (\ref{optim3}) using the 80 Million Tiny Images as $D_{out}^{OE}$. During fine-tuning, we use the SGD algorithm with momentum 0.9 and a cosine learning rate \citep{loshchilov-ICLR17SGDR} with an initial value 0.001 using a batch size of 128 for data sampled from $D_{in}$ and a batch size of 256 for data sampled from $D_{out}^{OE}$. For CIFAR-10 and CIFAR-100 experiments, we fine-tune the network for 30 and 20 epochs respectively, while for SVHN the corresponding number of epochs was 5. The values of the hyper-parameters $\lambda_1$ and $\lambda_2$ were chosen in the range $[0.03,0.12]$ using a separate validation dataset consisting of both in- and out-of-distribution images similar to \citet{Lee:2018:SUF:3327757.3327819}. The validation dataset is described in \ref{post_val_image}. The results are shown in Table~\ref{Mahalanobis}.

\begin{table*}[h]
\begin{adjustbox}{max width=\textwidth}
\begin{tabular}{cc|cc|cc|cc|cc|cc}
\multicolumn{2}{c}{}&\multicolumn{2}{c}{TNR95$\uparrow$}&\multicolumn{2}{c}{AUROC$\uparrow$}&\multicolumn{2}{c}{DAcc$\uparrow$}&\multicolumn{2}{c}{AUPRin$\uparrow$}&\multicolumn{2}{c}{AUPRout$\uparrow$}\\
\cline{3-12} 
${D}_{in}$&${D}_{out}^{test}$&MD&OECC+MD&MD&OECC+MD&MD&OECC+MD&MD&OECC+MD&MD&OECC+MD\\
\hline
\multirow{3}{*}{{{SVHN}}}&CIFAR-10&98.4&\textbf{99.9}&99.3&\textbf{99.9}&96.9&\textbf{99.2}&99.7&\textbf{100.0}&97.0&\textbf{99.6}\\
&TinyImageNet&99.9&\textbf{100.0}&99.9&\textbf{100.0}&99.1&\textbf{99.9}&99.9&\textbf{100.0}&99.1&\textbf{100.0}\\
&LSUN&99.9&\textbf{100.0}&99.9&\textbf{100.0}&99.5&\textbf{100.0}&99.9&\textbf{100.0}&99.1&\textbf{100.0}\\
\hline
\multirow{3}{*}{{{CIFAR-10}}}&SVHN&96.4&\textbf{97.3}&99.1&\textbf{99.2}&95.8&\textbf{96.3}&98.3&\textbf{98.4}&99.6&99.6\\
&TinyImageNet&97.1&\textbf{98.8}&99.5&\textbf{99.6}&96.3&\textbf{97.3}&\textbf{99.5}&99.4&99.5&\textbf{99.6}\\
&LSUN&98.9&\textbf{99.7}&99.7&\textbf{99.8}&97.7&\textbf{98.5}&\textbf{99.7}&99.5&99.7&\textbf{99.8}\\
\hline
\multirow{3}{*}{{{CIFAR-100}}}&SVHN&91.9&\textbf{93.0}&98.4&\textbf{98.7}&93.7&\textbf{94.2}&96.4&\textbf{97.1}&99.3&\textbf{99.5}\\
&TinyImageNet&90.9&\textbf{92.3}&98.2&\textbf{98.3}&93.3&\textbf{93.9}&98.2&\textbf{98.3}&98.2&\textbf{98.3}\\
&LSUN&90.9&\textbf{95.6}&98.2&\textbf{98.6}&93.5&\textbf{95.4}&98.4&98.4&97.8&\textbf{98.3}\\
\hline
\end{tabular}
\end{adjustbox}
\caption{\label{Mahalanobis}Comparison between the Mahalanobis Detector (MD) \citep{Lee:2018:SUF:3327757.3327819} and the combination of OECC with the MD method using a ResNet-34 architecture. Similar to \cite{Lee:2018:SUF:3327757.3327819}, the hyper-parameters are tuned using a validation dataset of in- and out-of-distribution data as described in \ref{post_val_image}.}
\end{table*}

\subsubsection{A Combination of OECC and Gram Matrices Method for OOD Detection}
Recently, \citet{ch2019detecting} proposed a post-training method for OOD detection that does not require access to OOD data for hyper-parameter tuning, unlike the MD method \citep{Lee:2018:SUF:3327757.3327819}. More specifically, they proposed the use of higher order Gram matrices to compute pairwise feature correlations between the channels of each layer of a DNN. Subsequently, after computing the minimum and maximum values of the correlations for every class $c$ in which an example generated by $D_{in}$ is classified, they used those values to calculate the layerwise deviation of each test sample, i.e. the deviation of test sample from the images seen during training with respect to each of the layers. Finally, they calculated the total deviation by taking a normalized sum of the layerwise deviations and using a threshold $\tau$, they classified a sample as OOD if its corresponding total deviation was above the threshold. The experimental results presented in \citet{ch2019detecting} showed that GM method outperforms MD method in most of the experiments without requiring access to OOD samples to tune its parameters. However, it should be noted that GM, in its current form, does not perform equally well when the samples from $D_{out}^{test}$ are close to $D_{in}$, as it happens for instance in the case where CIFAR-10 is used as $D_{in}$ and CIFAR-100 is used as $D_{out}^{test}$.

\paragraph{ResNet experiments.}For the results related to the GM method, we initially trained the ResNet model using exactly the same training details presented in Section~\ref{section_4_2_2} and then we applied the GM method where the tuning of the normalizing factor used to calculate the total deviation of a test image is done using a randomly selected validation partition from $D_{in}^{test}$, as described in \citet{ch2019detecting}. For the combined OECC$+$GM method, we initially trained the ResNet model as described above, then we fine-tuned it using the loss function described by (\ref{optim3}) and finally, we applied the GM method. During fine-tuning, we used the SGD algorithm with momentum 0.9 and a cosine learning rate \citep{loshchilov-ICLR17SGDR} with an initial value 0.001 using a batch size of 128 for data sampled from $D_{in}$ and a batch size of 256 for data sampled from $D_{out}^{OE}$. In our experiments, the 80 Million Tiny Images dataset \citep{Torralba:2008:MTI:1444381.1444403} was considered as $D_{out}^{OE}$. For CIFAR-10 experiments, we fine-tuned the network for 30 epochs, for CIFAR-100 we fine-tuned it for 10, while for SVHN the corresponding number of epochs was 5. Note that in the previous experiment, in which we combined the OECC method with the MD method, the hyper-parameters $\lambda_1$ and $\lambda_2$ of (\ref{optim3}) were tuned using a validation dataset of in- and out-of-distribution data as described in \ref{post_val_image} since this is required by the MD method. On the contrary, in this experiment, the hyper-parameters $\lambda_1$ and $\lambda_2$ of (\ref{optim3}) were tuned using a separate validation dataset $D_{out}^{val}$ described in \ref{post_val_image}. Note that $D_{out}^{val}$ and $D_{out}^{test}$ are disjoint. Therefore, for these experiments, no access to $D_{out}^{test}$ was assumed. The results of the experiments are shown in Table~\ref{FCorr}. The values of the hyper-parameters $\lambda_1$ and $\lambda_2$ were chosen in the range $[0.03,0.12]$.

\begin{table*}[h]
\begin{adjustbox}{max width=\textwidth}
\begin{tabular}{cc|cc|cc|cc|cc|cc}
\multicolumn{2}{c}{}&\multicolumn{2}{c}{TNR95$\uparrow$}&\multicolumn{2}{c}{AUROC$\uparrow$}&\multicolumn{2}{c}{DAcc$\uparrow$}&\multicolumn{2}{c}{AUPRin$\uparrow$}&\multicolumn{2}{c}{AUPRout$\uparrow$}\\
\cline{3-12} 
${D}_{in}$&${D}_{out}^{test}$&GM&OECC+GM&GM&OECC+GM&GM&OECC+GM&GM&OECC+GM&GM&OECC+GM\\
\hline
\multirow{3}{*}{{{SVHN}}}&CIFAR-10&85.8&\textbf{98.3}&97.3&\textbf{99.3}&92.0&\textbf{96.9}&98.9&\textbf{99.7}&93.2&\textbf{97.9}\\
&TinyImageNet&99.3&\textbf{100.0}&99.7&\textbf{100.0}&97.9&\textbf{99.5}&99.9&\textbf{100.0}&99.3&\textbf{99.9}\\
&LSUN&99.6&\textbf{100.0}&99.8&\textbf{100.0}&98.5&\textbf{99.8}&99.9&\textbf{100.0}&99.5&\textbf{100.0}\\
\hline
\multirow{3}{*}{{{CIFAR-10}}}&SVHN&97.6&\textbf{99.2}&99.5&\textbf{99.7}&96.7&\textbf{98.0}&98.4&\textbf{99.3}&99.8&\textbf{99.9}\\
&TinyImageNet&98.7&\textbf{99.6}&99.7&\textbf{99.8}&97.8&\textbf{98.3}&99.6&\textbf{99.8}&\textbf{99.8}&99.7\\
&LSUN&99.6&\textbf{99.9}&99.8&\textbf{99.9}&98.6&\textbf{99.0}&99.9&99.9&\textbf{99.9}&99.8\\
\hline
\multirow{3}{*}{{{CIFAR-100}}}&SVHN&80.8&\textbf{87.2}&96.0&\textbf{97.1}&89.6&\textbf{91.9}&90.5&\textbf{93.0}&98.5&\textbf{98.7}\\
&TinyImageNet&94.8&\textbf{95.8}&\textbf{98.9}&98.8&95.0&\textbf{95.5}&98.8&98.8&\textbf{99.0}&98.5\\
&LSUN&96.6&\textbf{98.2}&99.2&\textbf{99.3}&96.0&\textbf{96.8}&99.2&\textbf{99.4}&\textbf{99.2}&99.0\\
\hline
\end{tabular}
\end{adjustbox}
\caption{\label{FCorr}Comparison between the Gramian Matrices (GM) method \citep{ch2019detecting} versus the combination of OECC method and the GM method using a ResNet-34 architecture. The tuning of the hyperparameters $\lambda_1$ and $\lambda_2$ of (\ref{optim3}) is done using a separate validation dataset $D_{out}^{val}$ described in \ref{post_val_image}. Note that $D_{out}^{val}$ and $D_{out}^{test}$ are disjoint.}
\end{table*}

\paragraph{DenseNet experiments} For the results related to the GM method, we used the pre-trained DenseNet \citep{huang2017densely} model provided by \citet{2017arXiv170602690L}. The network has depth $L=100$, growth rate $m=12$ and dropout rate 0. It has been trained using the stochastic gradient descent algorithm with Nesterov momentum \citep{Duchi:2011:ASM:1953048.2021068, kingma2014adam} for 300 epochs with batch size 64 and momentum 0.9. The learning rate started at 0.1 and was dropped by a factor of 10 at $50\%$ and $75\%$ of the training progress, respectively. Subsequently, we applied the GM method \citep{ch2019detecting} where the tuning of the normalizing factor used to calculate the total deviation of a test image was done using a randomly selected validation partition from $D_{in}^{test}$ as described in \citet{ch2019detecting}. For the combined OECC$+$GM method, we fine-tuned the pre-trained DenseNet network model provided by \citet{2017arXiv170602690L} using the OECC loss function described by (\ref{optim3}) and then we applied the GM method. During fine-tuning, we used the SGD algorithm with momentum 0.9 and a cosine learning rate \citep{loshchilov-ICLR17SGDR} with an initial value 0.001 for CIFAR-10 and SVHN experiments and 0.01 for the CIFAR-100 experiments using a batch size of 128 for data sampled from $D_{in}$ and a batch size of 256 for data sampled from $D_{out}^{OE}$. In our experiments, the 80 Million Tiny Images dataset \citep{Torralba:2008:MTI:1444381.1444403} was considered as $D_{out}^{OE}$. The DenseNet model was fine-tuned for 15 epochs for the CIFAR-10 experiments, for 10 epochs for the CIFAR-100 experiments, while for SVHN the corresponding number of epochs was 5. The hyperparameters $\lambda_1$ and $\lambda_2$ of the OECC method were tuned using a separate validation dataset $D_{out}^{val}$ described in \ref{post_val_image}. Note that $D_{out}^{val}$ and $D_{out}^{test}$ are disjoint. The experimental results are presented in Table~\ref{FCorr_Densenet}. The values of the hyper-parameters $\lambda_1$ and $\lambda_2$ were chosen in the range $[0.03,0.12]$.

\begin{table*}[h]
\begin{adjustbox}{max width=\textwidth}
\begin{tabular}{cc|cc|cc|cc|cc|cc}
\multicolumn{2}{c}{}&\multicolumn{2}{c}{TNR95$\uparrow$}&\multicolumn{2}{c}{AUROC$\uparrow$}&\multicolumn{2}{c}{DAcc$\uparrow$}&\multicolumn{2}{c}{AUPRin$\uparrow$}&\multicolumn{2}{c}{AUPRout$\uparrow$}\\
\cline{3-12} 
${D}_{in}$&${D}_{out}^{test}$&GM&OECC+GM&GM&OECC+GM&GM&OECC+GM&GM&OECC+GM&GM&OECC+GM\\
\hline
\multirow{3}{*}{{{SVHN}}}&CIFAR-10&80.4&\textbf{98.5}&95.5&\textbf{99.6}&89.1&\textbf{97.5}&89.6&\textbf{98.6}&97.8&\textbf{99.8}\\
&TinyImageNet&99.1&\textbf{99.9}&99.7&\textbf{100.0}&97.9&\textbf{99.7}&99.3&\textbf{99.9}&99.9&\textbf{100.0}\\
&LSUN&99.5&\textbf{100.0}&99.8&\textbf{100.0}&98.6&\textbf{99.9}&99.5&\textbf{99.9}&99.9&\textbf{100.0}\\
\hline
\multirow{3}{*}{{{CIFAR-10}}}&SVHN&96.1&\textbf{98.5}&99.1&\textbf{99.6}&95.9&\textbf{97.4}&96.8&\textbf{98.6}&99.7&\textbf{99.9}\\
&TinyImageNet&98.8&\textbf{99.3}&99.7&\textbf{99.8}&97.9&\textbf{98.3}&99.6&\textbf{99.7}&99.8&99.8\\
&LSUN&99.5&\textbf{99.8}&99.9&99.9&98.6&\textbf{99.0}&99.9&99.9&99.9&99.9\\
\hline
\multirow{3}{*}{{{CIFAR-100}}}&SVHN&\textbf{89.3}&88.3&\textbf{97.3}&96.9&\textbf{92.4}&91.9&\textbf{91.7}&91.1&\textbf{99.1}&98.5\\
&TinyImageNet&95.7&\textbf{96.1}&99.0&99.0&95.5&\textbf{95.7}&98.8&\textbf{98.9}&\textbf{99.1}&98.4\\
&LSUN&97.2&\textbf{98.1}&99.3&99.3&96.4&\textbf{96.9}&99.3&\textbf{99.4}&\textbf{99.4}&98.8\\
\hline
\end{tabular}
\end{adjustbox}
\caption{\label{FCorr_Densenet}Comparison between the Gramian Matrices (GM) method \citep{ch2019detecting} versus the combination of OECC method and the GM method using a DenseNet-100 architecture. The tuning of the hyperparameters $\lambda_1$ and $\lambda_2$ of (\ref{optim3}) is done using a separate validation dataset $D_{out}^{val}$ described in \ref{post_val_image}. Note that $D_{out}^{val}$ and $D_{out}^{test}$ are disjoint.}
\end{table*}

\paragraph{Discussion.}The results presented in Table~\ref{Mahalanobis}, Table~\ref{FCorr}, and Table~\ref{FCorr_Densenet} demonstrate the superior performance that can be achieved when combining training and post-training methods for OOD detection. More specifically, the MD method \citep{Lee:2018:SUF:3327757.3327819} extracts the features from all layers of a pre-trained softmax neural classifier and then calculates the Mahalanobis distance-based confidence score. The GM method \citep{ch2019detecting} also extracts the features from a pre-trained softmax neural classifier and then computes higher order Gram matrices to subsequently calculate pairwise feature correlations between the channels of each layer of a DNN. As also mentioned earlier, both of these methods are post-training methods for OOD detection. On the other hand, the simulation results presented in Table~\ref{Image_OE_zipped_exp} and Table~\ref{NLP_OE_zipped_exp} showed that OECC, which belongs to the category of training methods for OOD detection, can teach the DNN to learn feature representations that can better distinguish in- and out-of-distribution data compared to the baseline method \citep{hendrycks17baseline} and the OE method \citep{hendrycks2019oe}. Therefore, by feeding a post-training method like the MD method \citep{Lee:2018:SUF:3327757.3327819} and the GM method \citep{ch2019detecting} with better feature representations, it is expected that one can achieve superior results in the OOD detection task as it is also validated by the experimental results in Table~\ref{Mahalanobis}, Table~\ref{FCorr}, and Table~\ref{FCorr_Densenet}.

\subsection{Calibration Experiments}
\citet{Guo:2017:CMN:3305381.3305518} discovered that deep neural networks are not well
calibrated. 
In their initial
\begin{wrapfigure}{r}{0.38\textwidth}
\includegraphics[width=1.0\linewidth]{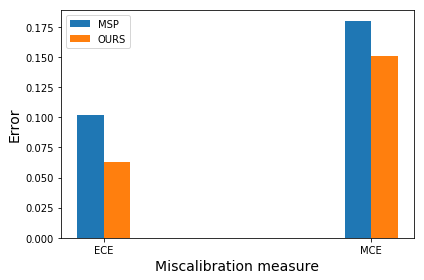} 
\includegraphics[width=1.0\linewidth]{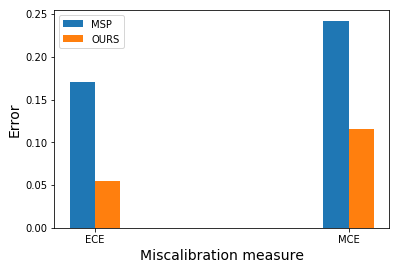} 
\caption{ECE and MCE for the MSP baseline detector and for the MSP baseline detector after fine-tuning with the OECC method described by (\ref{optim3}). {\it Top}: CIFAR-100. {\it Bottom}: SST.}
\label{fig:wrapfig2}
\end{wrapfigure}
experiment, they observed that a deep neural network like a 110-layer ResNet \citep{DBLP:journals/corr/HeZRS15} has an average confidence on its predictions for CIFAR-100 images that is much higher than its accuracy. 

As discussed earlier in Section~\ref{teleftaio} and as was also shown experimentally, the purpose of the second term of the loss function described by (\ref{optim3}) is to further distinguish in- and out-of-distribution examples and enhance the OOD detection capability of a neural network by pushing the maximum prediction probabilities produced by the softmax layer for the in-distribution examples towards the training accuracy of the DNN. Motivated by the fact that overconfident predictions constitute a symptom of overfitting \citep{DBLP:journals/corr/SzegedyVISW15} and also by the results of the experiments of \citet{Guo:2017:CMN:3305381.3305518}, we expect that by minimizing the squared distance between the training accuracy of the DNN and the average confidence in its predictions for examples drawn from $D_{in}$, not only will the neural network have a higher OOD detection capability, but it will also be more calibrated.    

To validate our hypothesis, we use two miscalibration measures, namely the Expected Calibration Error (ECE) and the Maximum Calibration Error (MCE) \citep{Naeini:2015:OWC:2888116.2888120}. ECE measures the difference in expectation between confidence and accuracy while MCE measures the worst-case deviation between confidence and accuracy. 

To calculate ECE, we first calculate the average confidence of the DNN for a selected number of examples $x_i$ sampled from $D_{in}^{test}$ and then we partition the examples into equally-spaced bins $\{B_m\}_{m=1}^M$ based on the output confidence of the DNN. Then, ECE is given by the following equation:
\begin{equation}
    ECE = \sum_{m=1}^{M} \frac{|B_m|}{n} \Big |acc(B_m) - conf(B_m) \Big |
\end{equation}
where $|B_m|$ is the number of examples in that bin, $n$ is the total number of examples, $acc(B_m)$ is the average classification accuracy of the DNN on the examples in $B_m$ and $conf(B_m)$ is the average confidence of the predictions made by the DNN for the examples in $B_m$. Using the same definitions, MCE is given by the following equation:
\begin{equation}
    MCE = \underset{m \in \{1,\ldots,M\}}{\text{max}} \Big |acc(B_m) - conf(B_m) \Big |
\end{equation}

To evaluate our method, we draw $n=1000$ test samples from $D_{in}$ and we compare the ECE and MCE miscalibration errors for the MSP baseline detector \citep{hendrycks17baseline} and the MSP detector fine-tuned with the OECC method described by (\ref{optim3}). We partition the data into $M=15$ bins. In Figure~\ref{fig:wrapfig2}, we plot the miscalibration errors considering as $D_{in}$ the CIFAR-100 and the SST datasets, respectively. For the CIFAR-100 experiments, we train a 40-2 WRN with exactly the same training details as in Section~\ref{refer_cal_1}. For the SST experiments, we train a 2-layer GRU with exactly the same training details as in Section~\ref{refer_cal_2}. As can be observed from the results of Figure~\ref{fig:wrapfig2}, the minimization of the squared distance between the average confidence of the neural network and its training accuracy through the second term of (\ref{optim3}) also generalizes to the test set by reducing the miscalibration errors for both image and text datasets.

\section{Conclusion}
In this paper, we proposed a novel method for OOD detection, called Outlier Exposure with Confidence Control (OECC). OECC includes two regularization terms the first of which minimizes the total variation distance between the output distribution of the softmax layer of a DNN and the uniform distribution, while the second minimizes the Euclidean distance between the training accuracy of a DNN and the average confidence in its predictions on the training set. Experimental results showed that the proposed method achieves superior results in OOD detection with OE \citep{hendrycks2019oe} in both image and text classification tasks. Additionally, we experimentally showed that our method can be combined with state-of-the-art post-training methods for OOD detection like the Mahalanobis Detector (MD) \citep{Lee:2018:SUF:3327757.3327819} and the Gramian Matrices method (GM) \citep{ch2019detecting} demonstrating the desirability of combination of training and post-training methods for OOD detection in the future research efforts.  

\clearpage

\section*{Acknowledgments}
We thank Google for donating Google Cloud Platform research credits used in this research. 


\bibliography{iclr2020_conference}
\bibliographystyle{iclr2020_conference}

\clearpage

\appendix
\section{Expanded Image OOD detection Results and Datasets used for Comparison with State-of-the-Art in OE}
\subsection{Image OOD detection Results}\label{image_expanded}
\begin{table}[h]
\begin{center}
\begin{tabular}{cl|cc|cc|cc}
\multicolumn{2}{c}{}&\multicolumn{2}{c}{FPR95$\downarrow$}&\multicolumn{2}{c}{AUROC$\uparrow$}&\multicolumn{2}{c}{AUPR$\uparrow$}\\
\cline{3-8} 
${D}_{in}$&${D}_{out}^{test}$&+OE&OECC&+OE&OECC&+OE&OECC\\
\hline
\multirow{8}{*}{{\rotatebox[origin=c]{90}{SVHN}}}&Gaussian&0.0&0.0&100.&100.&100.&99.4\\
&Bernulli&0.0&0.0&100.&100.&100.&99.2\\
&Blobs   &0.0&0.0&100.&100.&100.&99.6\\
&Icons-50&0.3&0.1&99.8&99.9&99.2&99.5\\
&Textures&0.2&0.1&100.&100.&99.7&99.6\\
&Places365&0.1&0.0&100.&100.&99.9&99.7\\
&LSUN    &0.1&0.0&100.&100.&99.9&99.7\\
&CIFAR-10&0.1&0.0&100.&100.&99.9&99.7\\
\hline
&Mean&0.10&\textbf{0.03}&99.98&\textbf{99.99}&\textbf{99.83}&99.55\\
\toprule[1.5pt]
\multirow{8}{*}{{\rotatebox[origin=c]{90}{CIFAR-10}}}&Gaussian&0.7&0.7&99.6&99.8&94.3&99.0\\
&Rademacher   &0.5&1.1&99.8&99.6&97.4&97.6\\
&Blobs        &0.6&1.5&99.8&99.1&98.9&91.7\\
&Textures     &12.2&4.0&97.7&98.9&91.0&95.0\\
&SVHN         &4.8&1.4&98.4&99.6&89.4&97.9\\
&Places365     &17.3&13.3&96.2&96.9&87.3&89.5\\
&LSUN         &12.1&6.7&97.6&98.4&89.4&91.9\\
&CIFAR-100    &28.0&23.8&93.3&94.9&76.2&82.0\\
\hline
&Mean&9.50&\textbf{6.56}&97.81&\textbf{98.40}&90.48&\textbf{93.08}\\
\midrule[1.5pt]
\multirow{8}{*}{{\rotatebox[origin=c]{90}{CIFAR-100}}}&Gaussian&12.1&0.7&95.7&99.7&71.1&97.2\\
&Rademacher&17.1&0.7&93.0&99.7&56.9&96.2\\
&Blobs     &12.1&1.3&97.2&99.6&86.2&96.3\\
&Textures  &54.4&50.1&84.8&87.8&56.3&61.5\\
&SVHN      &42.9&16.7&86.9&94.9&52.9&74.1\\
&Places365  &49.8&47.8&86.5&88.1&57.9&58.5\\
&LSUN      &57.5&56.6&83.4&85.9&51.4&53.0\\
&CIFAR-10  &62.1&57.2&75.7&78.7&32.6&35.2\\
\hline
&Mean&38.50&\textbf{28.89}&87.89&\textbf{91.80}&58.15&\textbf{71.50}\\
\bottomrule[1.5pt]
\end{tabular}
\end{center}
\caption{\label{image_table}Image OOD example detection for the maximum softmax probability (MSP) baseline detector after fine-tuning with OE \citep{hendrycks2019oe} versus fine-tuning with OECC given by (\ref{optim3}). All results are percentages and averaged over 10 runs. Values are rounded to the first decimal digit. As also mentioned before, for these results, no access to OOD samples was assumed.}
\end{table}

\subsection{$D_{in}$, $D_{out}^{OE}$ and $D_{out}^{test}$ for Image Experiments}\label{imagedata}
\textbf{SVHN:} The Street View House Number (SVHN) dataset \citep{37648} consists of $32\times32$ color images out of which 604,388 are used for training and 26,032 are used for testing. The dataset has 10 classes and was collected from real Google Street View images. Similar to \citet{hendrycks2019oe}, we rescale the pixels of the images to be in $[0,1]$.

\vspace{-5pt}
\textbf{CIFAR 10:} This dataset \citep{Krizhevsky09learningmultiple} contains 10 classes and consists of 60,000 $32\times32$ color images out of which 50,000 belong to the training and 10,000 belong to the test set. Before training, we standardize the images per channel similar to \citet{hendrycks2019oe}.
\vspace{-5pt}

\textbf{CIFAR 100:} This dataset \citep{Krizhevsky09learningmultiple} consists of 20 distinct superclasses each of which contains 5 different classes giving us a total of 100 classes. The total number of images in the dataset are 60,000 and we use the standard 50,000/10,000 train/test split. Before training, we standardize the images per channel similar to \citet{hendrycks2019oe}.
\vspace{-5pt}

\textbf{80 Million Tiny Images: }The 80 Million Tiny Images dataset \citep{Torralba:2008:MTI:1444381.1444403} was exclusively used in our experiments in order to represent $D^{OE}_{out}$. It consists of $32\times32$ color images collected from the Internet. Similar to \citet{hendrycks2019oe}, in order to make sure that $D^{OE}_{out}$ and $D^{test}_{out}$ are disjoint, we removed all the images of the dataset that appear on CIFAR 10 and CIFAR 100 datasets. 
\vspace{-15pt}

\textbf{Places365: }Places365 dataset introduced by \citet{7968387} was exclusively used in our experiments in order to represent $D^{test}_{out}$. It consists of millions of photographs of scenes.
\vspace{-5pt}

\textbf{Gaussian: }A synthetic image dataset created by i.i.d. sampling from an isotropic Gaussian distribution. 
\vspace{-5pt}

\textbf{Bernoulli: }A synthetic image dataset created by sampling from a Bernoulli distribution.
\vspace{-5pt}

\textbf{Blobs: }A synthetic dataset of images with definite edges.
\vspace{-5pt}

\textbf{Icons-50: }This dataset intoduced by \citet{2018arXiv180701697H} consists of 10,000 images belonging to 50 classes of icons. As part of preprocessing, we removed the class ``Number" in order to make it disjoint from the SVHN dataset.
\vspace{-5pt}

\textbf{Textures: }This dataset contains 5,640 textural images \citep{Cimpoi:2014:DTW:2679600.2680059}.
\vspace{-5pt}

\textbf{LSUN: }It consists of around 1 million large-scale images of scenes \citep{yu15lsun}.
\vspace{-5pt}

\textbf{Rademacher: }A synthetic image dataset created by sampling from a symmetric Rademacher distribution.

\subsection{Validation Data for Image Experiments}\label{val_image}
\textbf{Uniform Noise:} A synthetic image dataset where each pixel is sampled from $\mathcal{U}[0,1]$ or $\mathcal{U}[-1,1]$ depending on the input space of the classifier.
\vspace{-5pt}

\textbf{Arithmetic Mean: }A synthetic image dataset created by randomly sampling a pair of in-distribution images and subsequently taking their pixelwise arithmetic mean.
\vspace{-5pt}

\textbf{Geometric Mean: }A synthetic image dataset created by randomly sampling a pair of in-distribution images and subsequently taking their pixelwise geometric mean.
\vspace{-5pt}

\textbf{Jigsaw: }A synthetic image dataset created by partitioning an image sampled from $D_{in}$ into 16 equally sized patches and by subsequently permuting those patches.
\vspace{-5pt}

\textbf{Speckle Noised: }A synthetic image dataset created by applying speckle noise to images sampled from $D_{in}$.
\vspace{-5pt}

\textbf{Inverted Images: }A synthetic image dataset created by shifting and reordering the color channels of images sampled from $D_{in}$.
\vspace{-5pt}

\textbf{RGB Ghosted: }A synthetic image dataset created by inverting the color channels of images sampled from $D_{in}$.

\section{Expanded Text OOD detection Results and Datasets used for Comparison with State-of-the-Art in OE}
\subsection{$D_{in}$, $D_{out}^{OE}$ and $D_{out}^{test}$ for NLP Experiments}\label{nlpdata}
\textbf{20 Newsgroups: }This dataset contains 20 different newsgroups, each corresponding to a specific topic. It contains around 19,000 examples and we used the standard 60/40 train/test split.
\vspace{-5pt}

\textbf{TREC: }A question classification dataset containing around 6,000 examples from 50 different classes. Similar to \citet{hendrycks2019oe}, we used 500 examples for the test phase and the rest for training.
\vspace{-5pt}

\textbf{SST: }The Stanford Sentiment Treebank \citep{brusilovsky:socher2013recursive} is a binary classification dataset for sentiment prediction of movie reviews containing around 10,000 examples.
\vspace{-5pt}

\textbf{WikiText-2: }This dataset contains over 2 million articles from Wikipedia and is exclusively used as $D^{OE}_{out}$ in our experiments. We used the same preprocessing as in \citet{hendrycks2019oe} in order to have a valid comparison.
\vspace{-5pt}

\textbf{SNLI: }The Stanford Natural Language Inference (SNLI) corpus is a collection of 570,000 human-written English sentence pairs \citep{bowman2015large}. 
\vspace{-5pt}

\textbf{IMDB: }A sentiment classification dataset containing movies reviews.
\vspace{-5pt}

\textbf{Multi30K: } A dataset of English and German descriptions of images \citep{2016arXiv160500459E}. For our experiments, only the English descriptions were used.
\vspace{-5pt}

\textbf{WMT16: }A dataset used for machine translation tasks. For our experiments, only the English part of the test set was used.
\vspace{-5pt}

\textbf{Yelp: }A dataset containing reviews of users for businesses on Yelp.
\vspace{-5pt}

\textbf{EWT: }The English Web Treebank (EWT) consists of 5 different datasets: weblogs (EWT-W), newsgroups (EWT-N), emails (EWT-E), reviews (EWT-R) and questions-answers (EWT-A).

\subsection{Validation Data for NLP Experiments}\label{val_text}
The validation dataset $D_{out}^{val}$ used for the NLP OOD detection experiments was constructed as follows. For each $D_{in}$ dataset used, we used the rest two in-distribution datasets as $D_{out}^{val}$. For instance, during the experiments where {\it20 Newsgroups} represented $D_{in}$, we used {\it TREC} and {\it SST} as $D_{out}^{val}$ making sure that $D_{out}^{val}$ and $D_{out}^{test}$ are disjoint.

\subsection{Text OOD detection Results}\label{text_expanded}
\begin{table}[h]
\begin{center}
\begin{tabular}{cl|cc|cc|cc}
\multicolumn{2}{c}{}&\multicolumn{2}{c}{FPR90$\downarrow$}&\multicolumn{2}{c}{AUROC$\uparrow$}&\multicolumn{2}{c}{AUPR$\uparrow$}\\
\cline{3-8} 
${D}_{in}$&${D}_{out}^{test}$&+OE&OECC&+OE&OECC&+OE&OECC\\
\hline
\multirow{8}{*}{{\rotatebox[origin=c]{90}{20 Newsgroups}}}&SNLI&12.5&2.1&95.1&97.1&86.3&93.0\\
&IMDB&18.6&2.5&93.5&98.2&74.5&92.9\\
&Multi30K   &3.2&0.1&97.3&99.4&93.7&98.6\\
&WMT16&2.0&0.2&98.8&99.8&96.1&99.4\\
&Yelp&3.9&0.4&97.8&99.6&87.9&97.9\\
&EWT-A&1.2&0.2&99.2&99.8&97.3&98.4\\
&EWT-E    &1.4&0.1&99.2&99.9&97.2&98.9\\
&EWT-N&1.8&0.5&98.7&99.2&95.7&94.5\\
&EWT-R&1.7&0.1&98.9&99.4&96.6&98.3\\
&EWT-W&2.4&0.1&98.5&99.4&93.8&98.3\\
\hline
&Mean&4.86&\textbf{0.63}&97.71&\textbf{99.18}&91.91&\textbf{97.02}\\
\toprule[1.5pt]
\multirow{8}{*}{{\rotatebox[origin=c]{90}{TREC}}}&SNLI&4.2&0.8&98.1&99.1&91.6&94.9\\
&IMDB&0.6&0.6&99.4&98.9&97.8&97.1\\
&Multi30K   &0.3&0.2&99.7&99.9&99.0&99.6\\
&WMT16&0.2&0.2&99.8&99.9&99.4&99.6\\
&Yelp&0.4&0.8&99.7&99.1&96.1&92.9\\
&EWT-A&0.9&4.0&97.7&98.0&96.1&95.6\\
&EWT-E    &0.4&0.3&99.5&99.2&99.1&98.1\\
&EWT-N&0.3&0.2&99.6&99.9&99.2&99.6\\
&EWT-R&0.4&0.2&99.5&99.6&98.8&98.9\\
&EWT-W&0.2&0.2&99.7&99.6&99.4&98.9\\
\hline
&Mean&0.78&\textbf{0.75}&99.28&\textbf{99.32}&\textbf{97.64}&97.52\\
\midrule[1.5pt]
\multirow{8}{*}{{\rotatebox[origin=c]{90}{SST}}}&SNLI&33.4&7.4&86.8&95.8&52.0&76.4\\
&IMDB&32.6&10.8&85.9&95.8&51.5&77.6\\
&Multi30K   &33.0&5.1&88.3&97.9&58.9&86.9\\
&WMT16&17.1&3.6&92.9&98.3&68.8&88.1\\
&Yelp&11.3&15.6&92.7&95.2&60.0&81.1\\
&EWT-A&33.6&21.4&87.2&92.7&53.8&70.8\\
&EWT-E    &26.5&22.6&90.4&92.4&63.7&67.7\\
&EWT-N&27.2&19.2&90.1&93.6&62.0&67.4\\
&EWT-R&41.4&36.7&85.6&88.1&54.7&62.5\\
&EWT-W&17.2&36.7&92.8&88.1&66.9&62.5\\
\hline
&Mean&27.33&\textbf{17.91}&89.27&\textbf{93.79}&59.23&\textbf{74.10}\\
\bottomrule[1.5pt]
\end{tabular}
\end{center}
\caption{\label{NLP_table}NLP OOD example detection for the maximum softmax probability (MSP) baseline detector after fine-tuning with OE \citep{hendrycks2019oe} versus fine-tuning with OECC given by (\ref{optim3}). All results are percentages and the result of 10 runs. Values are rounded to the first decimal digit. As also mentioned before, for these results, no access to OOD samples was assumed.}
\end{table}

\section{Image Datasets used for Combination of OECC with Post-Training Methods}
\subsection{$D_{in}$, $D_{out}^{OE}$ and $D_{out}^{test}$ for Image Experiments}\label{post_image_data}

\textbf{SVHN:} In these experiments, we only used 73,257 training and 26,032 test images, i.e. in contrast with the experiments where we compared with the OE method, we did not use the extra SVHN dataset for training.  

\textbf{CIFAR-10:} Similar to the description in \ref{imagedata}.

\textbf{CIFAR-100:} Similar to the description in \ref{imagedata}.

\textbf{80 Million Tiny Images: }Similar to the description in \ref{imagedata}. 

\textbf{TinyImageNet: }This dataset consists of 10,000 test images with 200 image classes from a subset of ImageNet images. Similar to \cite{Lee:2018:SUF:3327757.3327819} and \cite{ch2019detecting}, we downsized the data to $32\times32$.

\textbf{LSUN: }In these experiments, we only used 10,000 test images of 10 different scenes. Similar to \cite{Lee:2018:SUF:3327757.3327819} and \cite{ch2019detecting}, we downsized the data to $32\times32$.

\subsection{Validation Data for Image Experiments}\label{post_val_image}
For the results of Table~\ref{Mahalanobis}, similar to \cite{Lee:2018:SUF:3327757.3327819}, the validation set consists of 1,000 images from each in- and out-of-distribution pair.

For the results of Table~\ref{FCorr} and Table~\ref{FCorr_Densenet}, the validation data used were similar to the ones described in \ref{val_image}. Note that $D^{val}_{out}$ and $D^{test}_{out}$ are disjoint.

\end{document}